\documentclass[11pt]{article}

\usepackage[a4paper,margin=1in]{geometry}
\usepackage{amsmath,amssymb}
\usepackage[nopatch=footnote]{microtype}
\usepackage{booktabs}
\usepackage{graphicx}
\usepackage{adjustbox}
\usepackage{pgfplots}
\usepackage{hyperref}
\pgfplotsset{compat=1.18}
\newcommand{\best}[1]{\ensuremath{\mathbf{#1}}}
\pgfmathdeclarefunction{normalcdf}{1}{%
	\pgfmathparse{0.5*(1+((#1)<0 ? -1 : 1)*(1-(((((1.061405429*(1/(1+0.3275911*abs(#1)/sqrt(2)))-1.453152027)*(1/(1+0.3275911*abs(#1)/sqrt(2)))+1.421413741)*(1/(1+0.3275911*abs(#1)/sqrt(2)))-0.284496736)*(1/(1+0.3275911*abs(#1)/sqrt(2)))+0.254829592)*(1/(1+0.3275911*abs(#1)/sqrt(2))))*exp(-(#1)^2/2)))}%
}

\title{A Structural Interpretation of GELU and Threshold-Transmission Activations via the First-Order Loss Function}
\author{
Roberto Rossi \\
  \small Business School, University of Edinburgh, Edinburgh, UK \\
  \small \texttt{roberto.rossi@ed.ac.uk}
}
\date{}

\begin{document}

\maketitle

\begin{abstract}
The Gaussian Error Linear Unit is usually motivated as the expected output of an input-dependent Bernoulli gate. This work gives an alternative interpretation: GELU is the expected output of a hard linear gate with a Gaussian random threshold. This view provides a generative interpretation for the Bernoulli gate: the gate opens once the input clears a latent Gaussian threshold. This interpretation stems from a decomposition based on well-known results in stochastic inventory theory and leads to a threshold-transmission family that includes ReLU, GELU, SiLU/Swish, and hard swish as special cases. By considering a latent uniform threshold, we recover a hard-swish-like piecewise-polynomial gate whose nonlinear transition is confined to a finite interval, yielding fixed- and learned-width variants. Controlled experiments on compact vision and language models show that calibrated or learned uniform-threshold gates are consistently competitive with GELU, ReLU, and SiLU/Swish, display architecture-dependent learned widths, and use the finite transition region nontrivially.
\end{abstract}

\section{Introduction}

The Gaussian Error Linear Unit \cite{hendrycks2016gelu} is a widely adopted activation function \cite{Dubey2022,kunc2024decadesactivationscomprehensivesurvey} defined as
$$
\operatorname{GELU}(z)=z\Phi(z),
$$
where $ \Phi $ denotes the cumulative distribution function of a standard normal random variable.
The authors in \cite{hendrycks2016gelu} define GELU as the expectation of a modification to Adaptive Dropout \cite{DBLP:conf/nips/BaF13}, thus relating it to stochastic regularisers \cite{JMLR:v15:srivastava14a}.
In particular, they motivate GELU through the following probabilistic view of a neuron's output. 
Let $ B(z) $ be a Bernoulli random variable with success probability $ \Phi(z) $, i.e.
$B(z)\sim \operatorname{Bernoulli}(\Phi(z))$.
If the input $ z $ is retained with probability $ \Phi(z) $ and dropped with probability $ 1-\Phi(z) $, then on average the output is
$$
\mathbb{E}[zB(z)]
=
z\mathbb{E}[B(z)]
=
z\Phi(z).
$$
Thus GELU can be interpreted as the expected output of an input-dependent Bernoulli gate.

This explanation is formally correct and gives a useful regularisation-based motivation. It also supplies a statistical reason for the Gaussian CDF, since neuron inputs are often approximately normal, especially in the presence of Batch Normalisation. However, it leaves open a different structural question: beyond being the average output of a stochastic gate, does the expression $ z\Phi(z) $ arise from a more general principle? 

We address this question by making the following contributions: 
\begin{itemize}
\item We identify GELU as the signal-transmission component $z\Phi(z)$ of the Gaussian complementary first-order loss function (see e.g. \cite{hadley1963inventory,Zipkin2000-di}), which is typically used in inventory control and finance to measure an expected surplus.
\item We explain why the remaining truncated-moment correction $\phi(z)$ is not retained in the activation function: removing it preserves the origin and suppresses strongly negative inputs, while retaining it gives an expected-surplus accounting quantity.
\item We provide a generative interpretation for the Bernoulli gate: $\Phi(z)$ is the probability that the input clears a Gaussian latent threshold; by considering alternative threshold distributions, we derive a threshold-transmission family $a_F(z)=zF(z)$ that includes ReLU, SiLU/Swish, and hard swish.
\item In particular, we recover hard swish \cite{howard2019mobilenetv3} by studying uniform-threshold gates, and introduce calibrated and learned-width variants (UELU and TUELU). 
\item Finally, we provide controlled empirical evidence on compact vision and language models, comparing these threshold-derived gates with standard baselines while reporting closed/transition/open region occupancy.
\end{itemize}
The remainder of this work is structured as follows. 
Section~\ref{sec:first-order-loss-functions} reviews first-order loss functions and identifies the Gaussian decomposition underlying GELU. 
Section~\ref{sec:signal-transmission} distinguishes signal transmission from loss accounting and compares GELU with the full and centred complementary loss functions. 
Section~\ref{sec:uncertain-threshold} develops our uncertain threshold interpretation. 
Sections~\ref{sec:alternative-activations} and~\ref{sec:relation-existing-activations} introduce threshold-transmission alternatives and relate them to existing gated activations. 
Section~\ref{sec:computational-study} reports the computational study, and Section~\ref{sec:discussion-conclusion} concludes.

\section{Gaussian first-order loss functions}
\label{sec:first-order-loss-functions}

Let $T$ be a random variable.
In stochastic inventory theory, financial option pricing, and related recourse problems in stochastic programming \cite{Birge2011}, one often encounters the so-called first-order loss function~(see e.g. \cite{Zipkin2000-di}, p. 447)
$$
L(z)=\mathbb{E}[(T-z)^+],
$$
where $(x)^+=\max\{x,0\}$, and its complementary function \cite{Rossi2014}
$$
\widehat{L}(z)=\mathbb{E}[(z-T)^+].
$$
For a standard normal random variable $Z$ --- i.e. $Z\sim N(0,1)$ ---
with density $ \phi $ and distribution function $ \Phi $
these functions have closed forms:
$$
L(z)=\phi(z)-z(1-\Phi(z)), \qquad \widehat{L}(z)=\phi(z)+z\Phi(z).
$$

The second identity is the key one. Indeed,
$$
\widehat{L}(z)
=
\mathbb{E}[(z-Z)^+]
=
\int_{-\infty}^{z}(z-t)\phi(t)\,dt.
$$
Expanding the integral gives
$$
\widehat{L}(z)
=
z\int_{-\infty}^{z}\phi(t)\,dt
-
\int_{-\infty}^{z}t\phi(t)\,dt.
$$
Since
$$
\int_{-\infty}^{z}\phi(t)\,dt=\Phi(z)
\qquad\text{and}\qquad
\int_{-\infty}^{z}t\phi(t)\,dt=-\phi(z),
$$
we obtain
$$
\widehat{L}(z)=z\Phi(z)+\phi(z).
$$
From this identity, which is prominent in quantitative finance and operations research, it follows that
$$
\operatorname{GELU}(z)=\widehat{L}(z)-\phi(z).
$$

GELU is thus embedded directly inside the complementary first-order loss function of a standard normal random variable. The important point is not merely the algebraic identity, but the decomposition it induces: $ \widehat{L}(z) $ contains both a boundary-gated signal term $z\Phi(z)$ and a truncated-moment correction term $\phi(z)$.

\section{Signal transmission versus loss accounting}
\label{sec:signal-transmission}

The two terms in the decomposition we have introduced have different roles. The term $z\Phi(z)$ is the probability-weighted signal: it is the value $z$ multiplied by the probability that a standard normal variable lies below $z$,
$$
\mathbb{E}[z\mathbf{1}_{\{Z\le z\}}]=z\Phi(z).
$$
It answers a forward-pass question: how much of the input signal is transmitted across a Gaussian uncertain boundary? The term $\phi(z)$ is different. It is a truncated-normal correction needed when computing expected surplus:
\begin{align*}
\mathbb{E}[(z-Z)^+]
&=
\mathbb{E}[(z-Z)\mathbf{1}_{\{Z\le z\}}]
\\&=
z\Phi(z)-\mathbb{E}[Z\mathbf{1}_{\{Z\le z\}}]
\\&=
z\Phi(z)+\phi(z).
\end{align*}
This expectation depends not only on the probability of crossing the boundary, but also on the conditional location of $Z$ inside the truncated region. For an activation function, the goal is not to compute expected surplus or slack; it is to transmit or suppress the input signal. Retaining $z\Phi(z)$ while removing $\phi(z)$ therefore keeps the Gaussian boundary-gated signal and discards the loss-accounting correction.

This observation also explains why the full complementary loss is not a natural substitute for GELU. At the origin, $\widehat{L}(0)=\phi(0)>0$, so a zero input would produce a positive output. By contrast, $\operatorname{GELU}(0)=0$, which is appropriate for an activation intended to preserve the origin.

Centering the complementary first-order loss at 0,
$$
\widetilde{L}(z)
=
\widehat{L}(z)-\widehat{L}(0)
=
\phi(z)-\phi(0)+z\Phi(z),
$$
fixes the origin since $\widetilde{L}(0)=0$, and yields an especially simple derivative:
$$
\widetilde{L}'(z)=\widehat{L}'(z)=\Phi(z).
$$
Thus $ \widetilde{L} $ is an activation whose gradient is exactly the Gaussian gate.
However, this comes at a cost. Since $\phi(z)\to 0$ and $z\Phi(z)\to 0$ as $z\to -\infty$,
$$
\lim_{z\to -\infty}\widetilde{L}(z)
=
-\phi(0).
$$
The centred loss preserves the origin but maps strongly negative inputs to a negative plateau, $-\phi(0)=-1/\sqrt{2\pi}$. 

More generally, if an activation has derivative exactly equal to the Gaussian gate, $a'(z)=\Phi(z)$, then $a(z)=\phi(z)+z\Phi(z)+C$ for some constant $ C $. Choosing $ C=0 $ gives $ \widehat{L} $, which suppresses the negative tail but does not preserve the origin. Choosing $ C=-\phi(0) $ gives $ \widetilde{L} $, which preserves the origin but has a negative plateau. No choice of the additive constant can simultaneously enforce $a(0)=0$ and $\lim_{z\to -\infty}a(z)=0$.

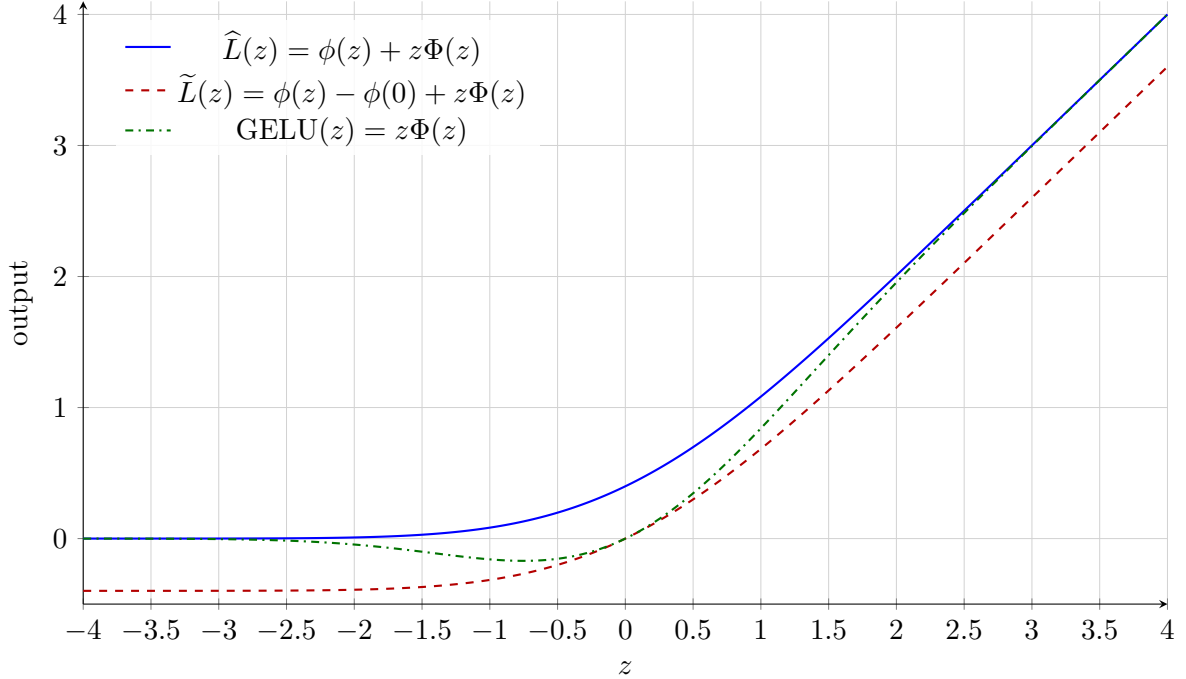
\begin{figure}
\centering
\begin{tikzpicture}
\begin{axis}[
	width=\columnwidth,
	height=0.6\columnwidth,
	xmin=-4, xmax=4,
	ymin=-0.5, ymax=4.1,
	axis lines=left,
	xlabel={$z$},
	ylabel={output},
	grid=both,
	grid style={line width=.1pt, draw=gray!20},
	major grid style={line width=.2pt, draw=gray!35},
	legend style={at={(0.03,0.97)}, anchor=north west, draw=none, fill=white, fill opacity=0.85, text opacity=1},
	samples=240,
	domain=-4:4,
]
\addplot[blue, thick] {1/sqrt(2*pi)*exp(-x^2/2) + x*normalcdf(x)};
\addlegendentry{$\widehat{L}(z)=\phi(z)+z\Phi(z)$}
\addplot[red!70!black, thick, dashed] {1/sqrt(2*pi)*exp(-x^2/2) - 1/sqrt(2*pi) + x*normalcdf(x)};
\addlegendentry{$\widetilde{L}(z)=\phi(z)-\phi(0)+z\Phi(z)$}
\addplot[green!45!black, thick, dash dot] {x*normalcdf(x)};
\addlegendentry{$\operatorname{GELU}(z)=z\Phi(z)$}
\end{axis}
\end{tikzpicture}
\caption{The complementary loss is positive at the origin, the centred complementary loss has a negative plateau, and GELU preserves both the origin and the negative-tail limit.}
\label{fig:comparison}
\end{figure}

GELU resolves this conflict by not being the antiderivative of the Gaussian gate. Instead, it gates the forward signal directly:
$$
\operatorname{GELU}(z)=z\Phi(z).
$$
Its derivative is therefore
$$
\frac{d}{dz}\operatorname{GELU}(z)
=
\Phi(z)+z\phi(z),
$$
not merely $ \Phi(z) $. The additional boundary term $ z\phi(z) $ is the price paid for obtaining the desired forward-pass behaviour. This term is negative on the negative side of the boundary, positive on the positive side, and vanishes in both tails, which explains the small non-monotone negative dip of GELU.
GELU is therefore ReLU-like in its positive-side linearity and negative-tail suppression, but it is not simply a monotone smoothing of ReLU. This comparison is summarised in Table~\ref{tab:comparison} and shown graphically in Figure~\ref{fig:comparison}.

\begin{table}[tbp]
\centering
\small
\begin{adjustbox}{max width=\columnwidth}
\begin{tabular}{@{}llll@{}}
\toprule
Function & Formula & Value at $ z=0 $ & Negative tail \\
\midrule
Complementary loss
&
$ \phi(z)+z\Phi(z) $
&
$ \phi(0)>0 $
&
$ 0 $
\\
Centred complementary loss
&
$ \phi(z)-\phi(0)+z\Phi(z) $
&
$ 0 $
&
$ -\phi(0) $
\\
GELU
&
$ z\Phi(z) $
&
$ 0 $
&
$ 0 $
\\
\bottomrule
\end{tabular}
\end{adjustbox}
\caption{The full complementary loss, its centred version, and GELU have different activation geometry. GELU preserves the origin and suppresses strongly negative inputs by retaining only the signal-transmission component.}
\label{tab:comparison}
\end{table}

To conclude, $\phi(z)$ is not an innocuous additive correction from the perspective of neural activations. It is essential for loss accounting, but retaining it changes the forward-pass activation geometry. Removing it converts a surplus-measuring object into a signal-transmission operator.

\section{Gaussian uncertain threshold interpretation}
\label{sec:uncertain-threshold}

The decomposition in the previous section suggests a complementary structural interpretation of GELU.

Suppose that a neuron transmits its input $z$ only if $z$ exceeds a random threshold $T$.
The corresponding hard gated output is $z\mathbf{1}_{\{T\le z\}}$. Let $T\sim N(0,1)$; 
taking expectation over $T$ gives
$$
\mathbb{E}[z\mathbf{1}_{\{T\le z\}}]
=
z\mathbb{P}(T\le z)
=
z\Phi(z).
$$
Hence, the average transmitted signal is
$$
\operatorname{GELU}(z)
=
\mathbb{E}[z\mathbf{1}_{\{T\le z\}}].
$$
This is closely related to the usual dropout-like interpretation, since for each fixed $z$, $ \mathbf{1}_{\{T\le z\}} $ is a Bernoulli random variable with success probability $ \Phi(z) $. The threshold view does not change the expectation. Its value is that it supplies a generative interpretation: the Bernoulli gate represents the event that {\em the input clears the uncertain threshold}, so $ \Phi(z) $ is also the probability of threshold clearance, complementing its role as an input-dependent keep probability.
Table~\ref{tab:gelu-statement-decomposition} in Appendix~\ref{app:symbols} summarises the mathematical objects connecting this threshold view to the expected-surplus decomposition.

Under this interpretation, ReLU \cite{nair2010rectified} corresponds to a deterministic threshold at zero:
$$
\operatorname{ReLU}(z)=z\mathbf{1}_{\{0\leq z\}}.
$$
GELU replaces this fixed threshold with a normally distributed uncertain threshold. It is therefore a smooth Gaussian relaxation of ReLU's hard threshold gate, although the resulting activation is not a globally monotone smoothing of ReLU itself.

The standard normal assumption is explicit in this view.
It should not be intended as requiring neural preactivations to be standard normal at every layer or stage of training.
Rather, standard GELU corresponds to expressing the uncertain boundary in centred, unit-scale coordinates.
A more general Gaussian threshold $T\sim N(\mu,\sigma^2)$ would produce
$$
a_{\mu,\sigma}(z)=z\Phi\!\left(\frac{z-\mu}{\sigma}\right).
$$
The standard normal case is therefore the normalised representative of this Gaussian-threshold family.

There is also a maximum-entropy interpretation \cite{jaynes03}.
Suppose that the standardised threshold error is known only to be real-valued, centred, and of unit variance.
Among all distributions on the real line satisfying these two moment constraints, the standard normal distribution has maximum entropy.
The Gaussian threshold is therefore the least informative centred unit-scale threshold law when no further information about support, skewness, or tail behaviour is imposed.
In this sense, standard GELU is not only the normalised Gaussian case but also the maximum-entropy threshold-transmission activation under the usual mean-variance normalisation.

\section{Exponential and uniform threshold laws}
\label{sec:alternative-activations}

The preceding construction is not limited to Gaussian uncertainty. Let $T$ be a threshold random variable with CDF $F$ and density $f$ --- for notational simplicity and without loss of generality we discuss the continuous case; discrete or mixed threshold laws are handled by replacing the density integral with the equivalent Lebesgue--Stieltjes expectation with respect to $F$.
The complementary first-order loss function associated with this threshold law is
$$
\widehat{L}_F(z)
=
\mathbb{E}[(z-T)^+]
=
\int_{-\infty}^{z}(z-u)f(u)\,du.
$$
It decomposes as
$$
\widehat{L}_F(z)
=
zF(z)+C_F(z),
\qquad
C_F(z)=-\int_{-\infty}^{z}u f(u)\,du,
$$
where $zF(z)$ is the transmitted signal and $C_F(z)$ is the truncated-moment correction. The activation associated with the threshold law keeps $zF(z)$ and discards $C_F(z)$.

Also in this case, the maximum-entropy viewpoint can be used to motivate alternative threshold laws. The Gaussian distribution is canonical when the threshold error is real-valued and only its mean and variance are specified. If the threshold is known to have bounded symmetric support, the maximum-entropy law on that support is uniform; if it is known to be nonnegative with a fixed mean, the maximum-entropy law is exponential. These alternatives therefore correspond to different information states about the latent threshold.

For example, an exponential threshold with rate $\lambda>0$ gives the one-sided activation
$$
\operatorname{ExpELU}_{\lambda}(z)
=
\begin{cases}
0, & z\le 0,\\
z(1-e^{-\lambda z}), & z>0.
\end{cases}
$$
It preserves the origin and the positive linear tail, but its one-sided support makes it closer in geometry to a smoothed ReLU than to the two-sided gates studied empirically below. 

The main alternative studied in this work is instead the {\em uniform threshold}. Let $ T $ be uniformly distributed on $ [-\beta,\beta] $, where $ \beta>0 $. Its distribution function is zero below $ -\beta $, increases linearly on $ [-\beta,\beta] $, and is one above $ \beta $. The corresponding uniform error linear unit is
$$
\operatorname{UELU}_{\beta}(z)
=
\begin{cases}
0, & z<-\beta,\\
\dfrac{z(z+\beta)}{2\beta}, & -\beta\le z\le \beta,\\
z, & z>\beta.
\end{cases}
$$
This activation replaces the smooth Gaussian transition by a finite-width deterministic ramp. Inputs below the support of the threshold distribution are suppressed, inputs above it are transmitted unchanged, and inputs inside the uncertain boundary layer are transmitted by a quadratic polynomial. Its derivative inside the boundary layer is the linear ramp
$$
\operatorname{UELU}_{\beta}'(z)
=
\frac{2z+\beta}{2\beta},
\qquad -\beta<z<\beta.
$$
Thus UELU corresponds to a threshold distribution with compact support: outside the interval $ [-\beta,\beta] $, the gate is fully closed or fully open. Inside the interval, the derivative is affine. The finite support creates kinks at the boundary points.

\begin{figure*}
\centering
\begin{minipage}{0.48\textwidth}
\centering
\includegraphics[width=\textwidth]{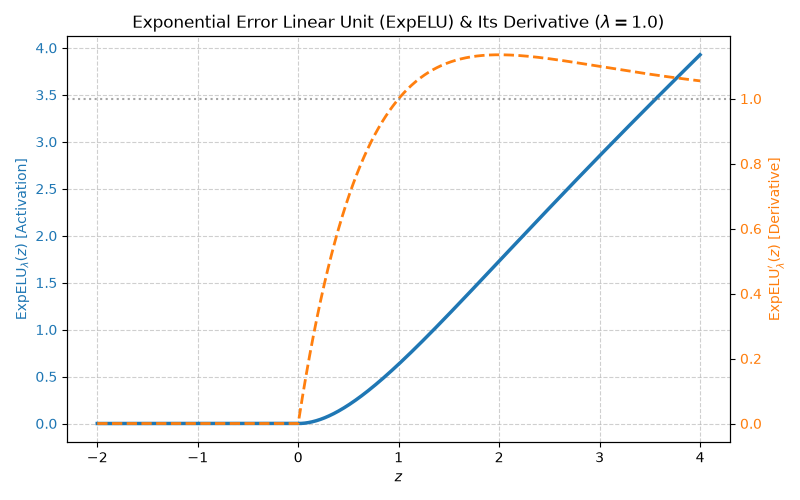}
\end{minipage}\hfill
\begin{minipage}{0.48\textwidth}
\centering
\includegraphics[width=\textwidth]{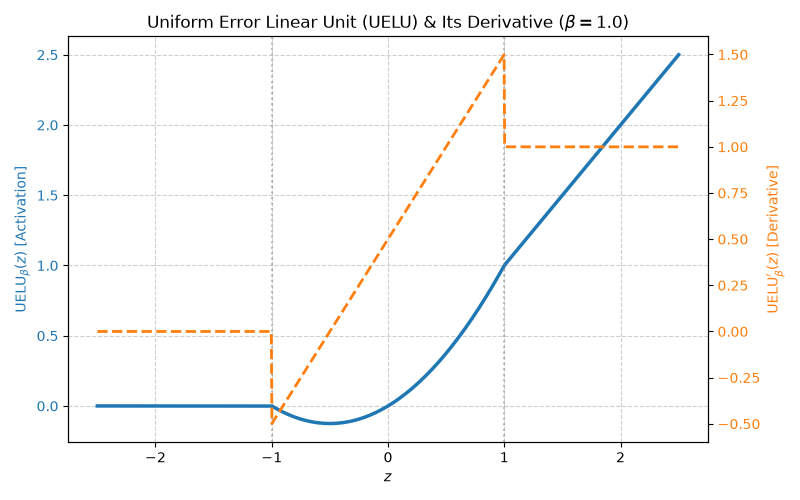}
\end{minipage}
\caption{Two threshold-transmission examples. Left: ExpELU for $\lambda=1$, illustrating one-sided threshold uncertainty as a contrast case. Right: UELU for $\beta=1$, the compact symmetric threshold law studied empirically below.}
\label{fig:uelu}
\end{figure*}

Note that UELU also has a small negative parabolic dip on $-\beta<z<0$, mirroring in piecewise-polynomial form the negative dip of GELU: boundary uncertainty removed from the forward loss-accounting term reappears as distinctive local geometry near the uncertain threshold. This behaviour is shown in Figure~\ref{fig:uelu} for $ \beta=1 $.

These examples show how the first-order-loss decomposition generates activation functions from different threshold laws. The detailed decompositions for the exponential and uniform cases are given in Appendix~\ref{app:alternative-activations}, together with a brief universal approximation note.

\section{Relation to existing gated activations}
\label{sec:relation-existing-activations}

The threshold-transmission construction can be leveraged to place the preceding formulas within the broader class of gated activations. Given a latent threshold distribution with CDF $F$, the transmitted signal has the common form
$$
a_F(z)=zF(z).
$$
Several relevant activation functions have this structure, as summarised in Table~\ref{tab:threshold-family}.

\begin{table*}[tbp]
\centering
\small
\begin{adjustbox}{max width=\textwidth}
\begin{tabular}{@{}lll@{}}
\toprule
Threshold law & Gate $F(z)$ & Activation $zF(z)$ \\
\midrule
Degenerate at zero & $\mathbf{1}\{z>0\}$ & ReLU \\
Gaussian & $\Phi(z)$ & GELU \\
Logistic & $\sigma(z)$ & SiLU/Swish \\
Uniform on $[-\beta,\beta]$ & $\operatorname{clip}\left((z+\beta)/(2\beta),0,1\right)$ & UELU / hard-swish family \\
Exponential on $[0,\infty)$ & $(1-e^{-\lambda z})\mathbf{1}\{z>0\}$ & Exponential-threshold unit \\
\bottomrule
\end{tabular}
\end{adjustbox}
\caption{Several rectified and gated activations can be written as threshold-transmission functions $zF(z)$ for different latent threshold laws.}
\label{tab:threshold-family}
\end{table*}

ReLU corresponds to a degenerate threshold at zero; GELU corresponds to a Gaussian threshold; and SiLU/Swish \cite{Elfwing2018}, $ z\sigma(z) $, is the logistic-threshold analogue of GELU: if $T$ has the standard logistic distribution, then $F(z)=\sigma(z)$ and the transmitted signal is $z\sigma(z)$. The corresponding complementary first-order loss is the softplus function, so SiLU/Swish is obtained by removing the logistic truncated-moment correction from softplus; the corresponding derivation is given in Appendix~\ref{app:alternative-activations}. 

This perspective also clarifies the relation between UELU and hard swish. The standard hard swish activation used in MobileNetV3-style architectures \cite{howard2019mobilenetv3} is commonly written as 
$$
z\operatorname{clip}\left(\frac{z+3}{6},0,1\right).
$$
Setting $\beta=3$ in UELU gives exactly this expression.
Thus the uniform-threshold derivation recovers hard swish and generalises its fixed transition width to an explicit uncertainty half-width $\beta$. The case $\beta=3$ corresponds to the conventional hard-swish scaling, while other values correspond to narrower or wider uniform threshold uncertainty.
In our computational study, SiLU/Swish serves as an established logistic-threshold baseline, and we focus on the uniform threshold case (i.e. UELU) because it has an exact connection to hard swish and exposes an explicit threshold-width parameter.

UELU and TUELU also relate to adaptive piecewise-quadratic activations. The Adaptive Quadratic Linear Unit (AQuLU) \cite{wu2023aqulu} has the form
$$
f(z_i)=
\begin{cases}
z_i, & z_i \ge (1-b_i)/a_i,\\
a_i z_i^2+b_i z_i, & -b_i/a_i\le z_i<(1-b_i)/a_i,\\
0, & z_i<-b_i/a_i,
\end{cases}
$$
with trainable parameters $a_i$ and $b_i$. 
UELU is the symmetric one-parameter specialisation $b_i=\tfrac{1}{2}$, $a_i=\tfrac{1}{2\beta}$: these choices give the breakpoints $-\beta$ and $\beta$, and the middle segment becomes
$$
a_i z^2+b_i z
=
\frac{z^2}{2\beta}+\frac{z}{2}
=
\frac{z(z+\beta)}{2\beta}.
$$
TUELU should therefore be interpreted as a restricted learned-width subfamily of AQuLU. Its restriction is deliberate: it ties the two AQuLU degrees of freedom to a single threshold-width parameter, preserving symmetric threshold support and giving $\beta$ a direct interpretation as boundary uncertainty.

Finally, we have not identified an exact prior formula corresponding to ExpELU in the activation-function survey \cite{kunc2024decadesactivationscomprehensivesurvey} or in the broader literature, though it is closely related in spirit to ReLU-like one-sided activations and exponential-family activations.

\section{Computational study}
\label{sec:computational-study}

The preceding sections give a structural interpretation of GELU and identify uniform-threshold members of the same threshold-transmission family. In this computational study we ask whether the explicit uniform threshold width is useful in compact architectures where GELU is a natural baseline, and whether the learned or calibrated width behaves consistently across model families. We compare fixed-width UELU, its trainable-width version TUELU, and standard activations. We also include a centred dynamic GELU variant,
$$
\operatorname{DGELU}(z,t)=z\Phi(z)+\gamma(t)\bigl(\phi(z)-\phi(0)\bigr),
$$
with $\gamma(t)$ annealed linearly from one to zero, as a direct test of whether retaining a centred remnant of the loss correction is useful during training. The experiments are designed to expose not only accuracy or perplexity, but also the operational use of the compact threshold region.

The first two experiments use the same CIFAR-100 split while changing the architecture from an MLP-Mixer \cite{NEURIPS2021_cba0a4ee} to a compact Vision Transformer \cite{dosovitskiy2021an}. The remaining three experiments test language modelling across multiple text settings: a small character-level GPT on Tiny Shakespeare, a token-level GPT on TinyStories, and a token-level GPT on WikiText-2. 
Throughout the study we compare standard GELU with fixed-width UELU,
$$
\operatorname{UELU}_{\beta}(z)=z\operatorname{clip}\left(\frac{z+\beta}{2\beta},0,1\right),
$$
for the calibrated width $\beta=\sqrt{\pi/2}\approx1.25$, and for the hard-swish width setting $\beta=3$. The value $\sqrt{\pi/2}$ follows from matching the local geometry of GELU and UELU at the origin. Near zero,
$$
\operatorname{GELU}(z)=z\Phi(z)\approx \frac{z}{2}+\phi(0)z^2,
$$
whereas inside the UELU transition region,
$$
\operatorname{UELU}_{\beta}(z)=\frac{z}{2}+\frac{z^2}{2\beta}.
$$
Equating the quadratic coefficients gives 
$$
1/(2\beta)=\phi(0)=1/\sqrt{2\pi}
$$ 
and hence $\beta=\sqrt{\pi/2}$. For this reason, in what follows we refer to this calibrated value of $\beta$ as GELU-matched. We also test TUELU, where a single positive $\beta$ is shared across all UELU layers and learned through a softplus parameterisation. In this latter setting, $\beta$ is once more initialised to the GELU-matched value $\sqrt{\pi/2}$. For UELU and TUELU, we record the fraction of preactivations in the closed region $z<-\beta$, transition region $-\beta\le z\le\beta$, and open region $z>\beta$. ReLU and SiLU/Swish \cite{Elfwing2018} serve as baselines.

\subsection{MLP-Mixer on CIFAR-100}

The first experiment uses a compact MLP-Mixer on CIFAR-100 \cite{krizhevsky2009learning}. We hold out 5,000 examples from the 50,000-example training set as validation data, normalise images with standard CIFAR-100 statistics, and use random cropping and horizontal flipping during training. The model uses $4\times4$ patches, 192-dimensional embeddings, and six Mixer blocks with token- and channel-mixing widths 96 and 384. It is trained for 20 epochs and five seeds with AdamW \cite{DBLP:conf/iclr/LoshchilovH19}, learning rate $3\cdot10^{-4}$, weight decay $0.05$, label smoothing \cite{7780677} $0.1$, batch size 128, and cosine decay.

\begin{figure*}
\centering
\begin{minipage}{0.48\textwidth}
\centering
\includegraphics[width=\textwidth]{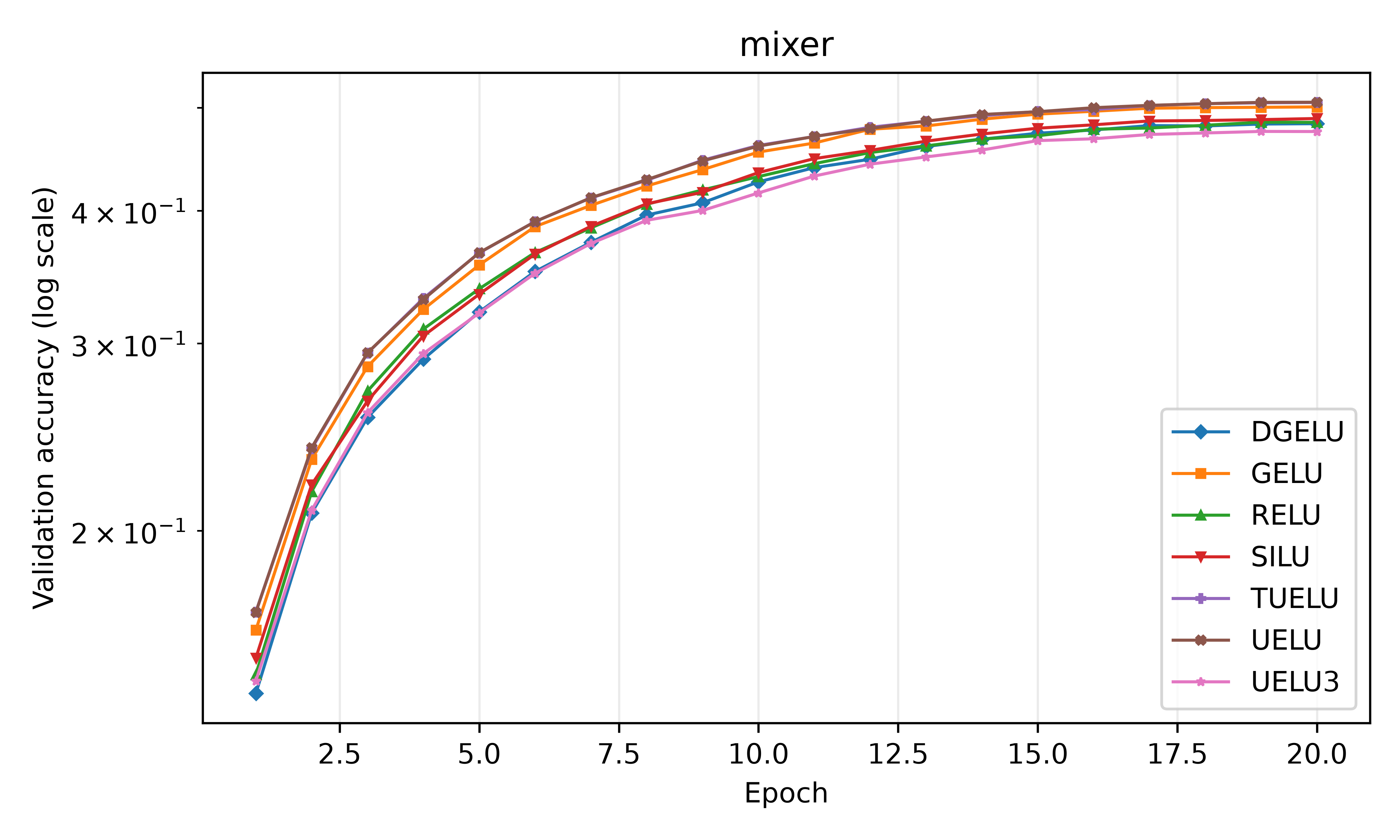}
\end{minipage}\hfill
\begin{minipage}{0.48\textwidth}
\centering
\includegraphics[width=\textwidth]{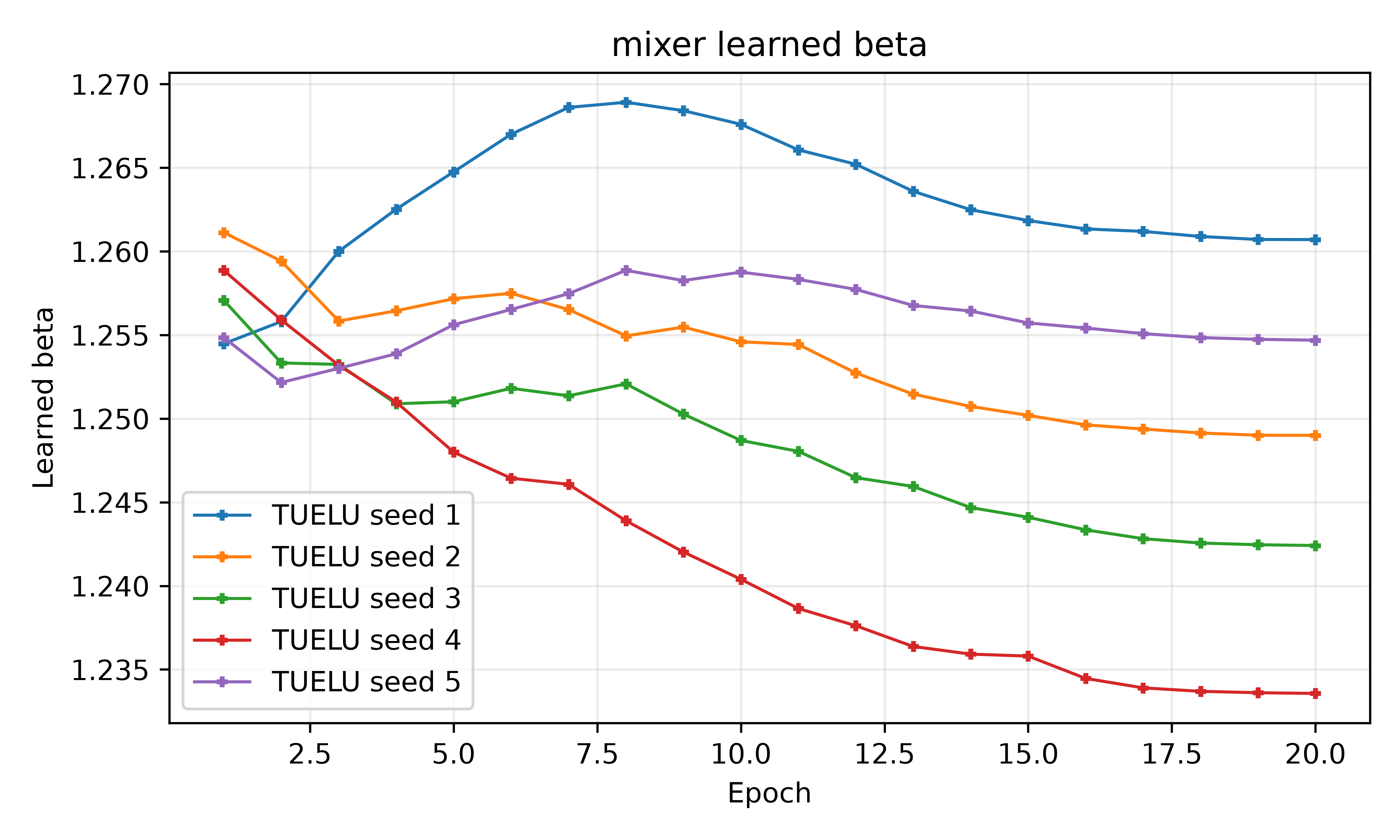}
\end{minipage}
\caption{MLP-Mixer results. Left: validation accuracy over training. Right: learned shared $\beta$ for TUELU.}
\label{fig:mixer-results}
\end{figure*}

The results are shown in Table~\ref{tab:mixer-results} and Figure~\ref{fig:mixer-results}. UELU with the GELU-matched width $\beta=1.25$ and TUELU both slightly improve mean test accuracy over GELU, ReLU, and SiLU; DGELU and Hard swish / UELU with $\beta=3$ underperform. TUELU learns $\beta\approx1.25$, remaining close to its GELU-matched initialisation and much narrower than the conventional hard-swish width.

\begin{table*}[tbp]
\centering
\resizebox{\textwidth}{!}{
\begin{tabular}{lllllll}
\toprule
Activation & $\beta$ & Learned $\beta$ & Val. acc. & Test acc. & Test loss & Regions C/T/O \\
\midrule
ReLU & -- & -- & $48.51\pm0.60$ & $49.15\pm0.41$ & $2.4180\pm0.0149$ & -- \\
SiLU & -- & -- & $48.86\pm1.04$ & $49.53\pm0.52$ & $2.4081\pm0.0182$ & -- \\
GELU & -- & -- & $50.14\pm0.69$ & $50.89\pm0.66$ & $2.3673\pm0.0187$ & -- \\
DGELU & -- & -- & $48.38\pm0.88$ & $49.01\pm0.78$ & $2.4304\pm0.0227$ & -- \\
Hard swish / UELU & $3.0$ & -- & $47.55\pm0.92$ & $48.32\pm0.58$ & $2.4490\pm0.0167$ & $0.55/99.3/0.18$ \\
UELU & $1.25$ & -- & \best{50.70\pm0.77} & $51.28\pm0.52$ & $2.3591\pm0.0163$ & $9.38/88.6/1.97$ \\
TUELU & $1.25$ & $1.248\pm0.009$ & $50.62\pm0.68$ & \best{51.29\pm0.45} & \best{2.3590\pm0.0159} & $9.43/88.6/1.98$ \\
\bottomrule
\end{tabular}
}
\caption{MLP-Mixer on CIFAR-100. Validation and test accuracies, test loss, and learned $\beta$ are reported as mean $\pm$ standard deviation over five seeds. The final column records mean percentages of UELU preactivations in the closed, transition, and open regions.}
\label{tab:mixer-results}
\end{table*}

\subsection{Compact Vision Transformer on CIFAR-100}

The second experiment uses a compact Vision Transformer on the same CIFAR-100 split and preprocessing pipeline. The model uses $4\times4$ patches, a learned class token, learned positional embeddings, 192-dimensional embeddings, six pre-normalisation Transformer blocks, six attention heads, and feedforward width 384. The optimiser and schedule match the Mixer experiment.

\begin{figure*}
\centering
\begin{minipage}{0.48\textwidth}
\centering
\includegraphics[width=\textwidth]{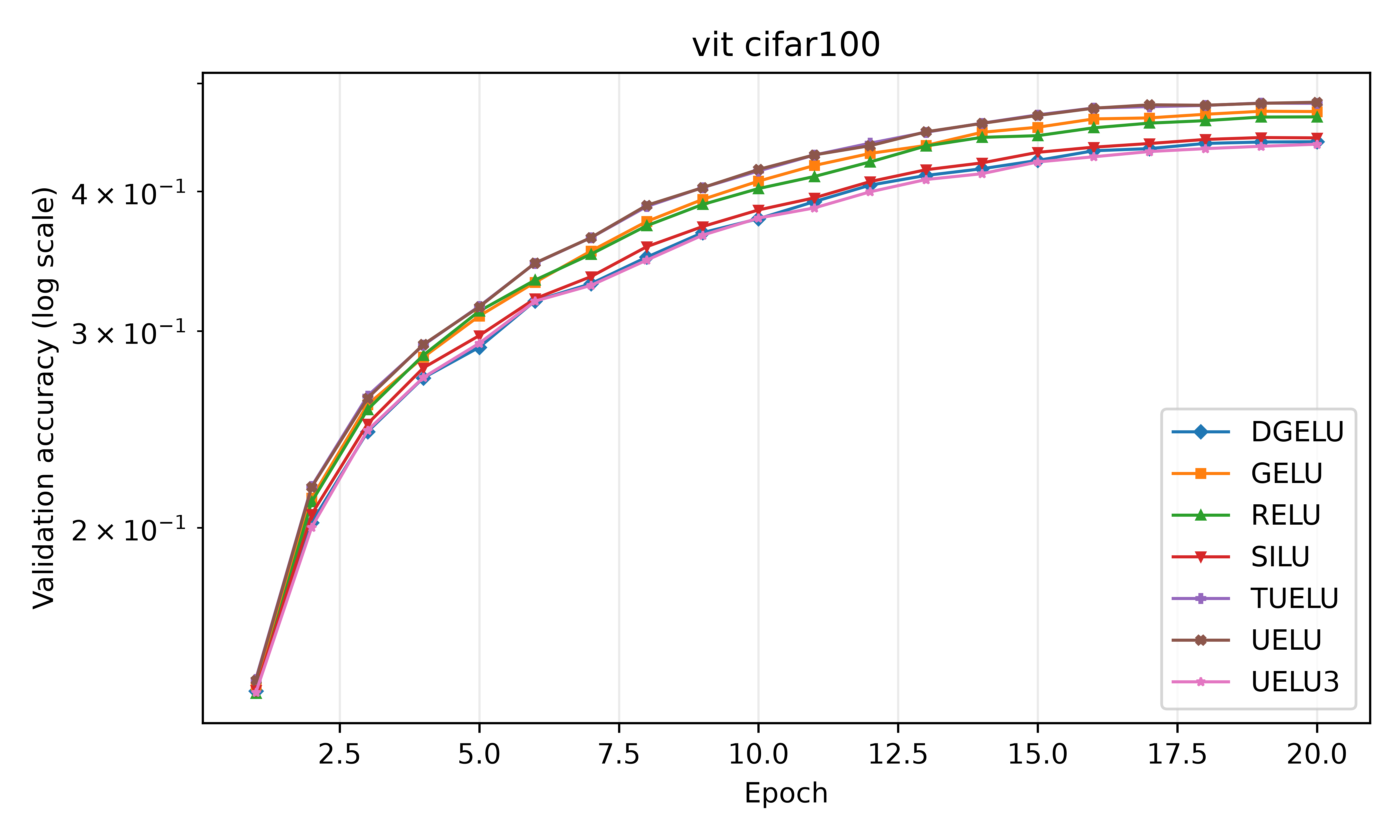}
\end{minipage}\hfill
\begin{minipage}{0.48\textwidth}
\centering
\includegraphics[width=\textwidth]{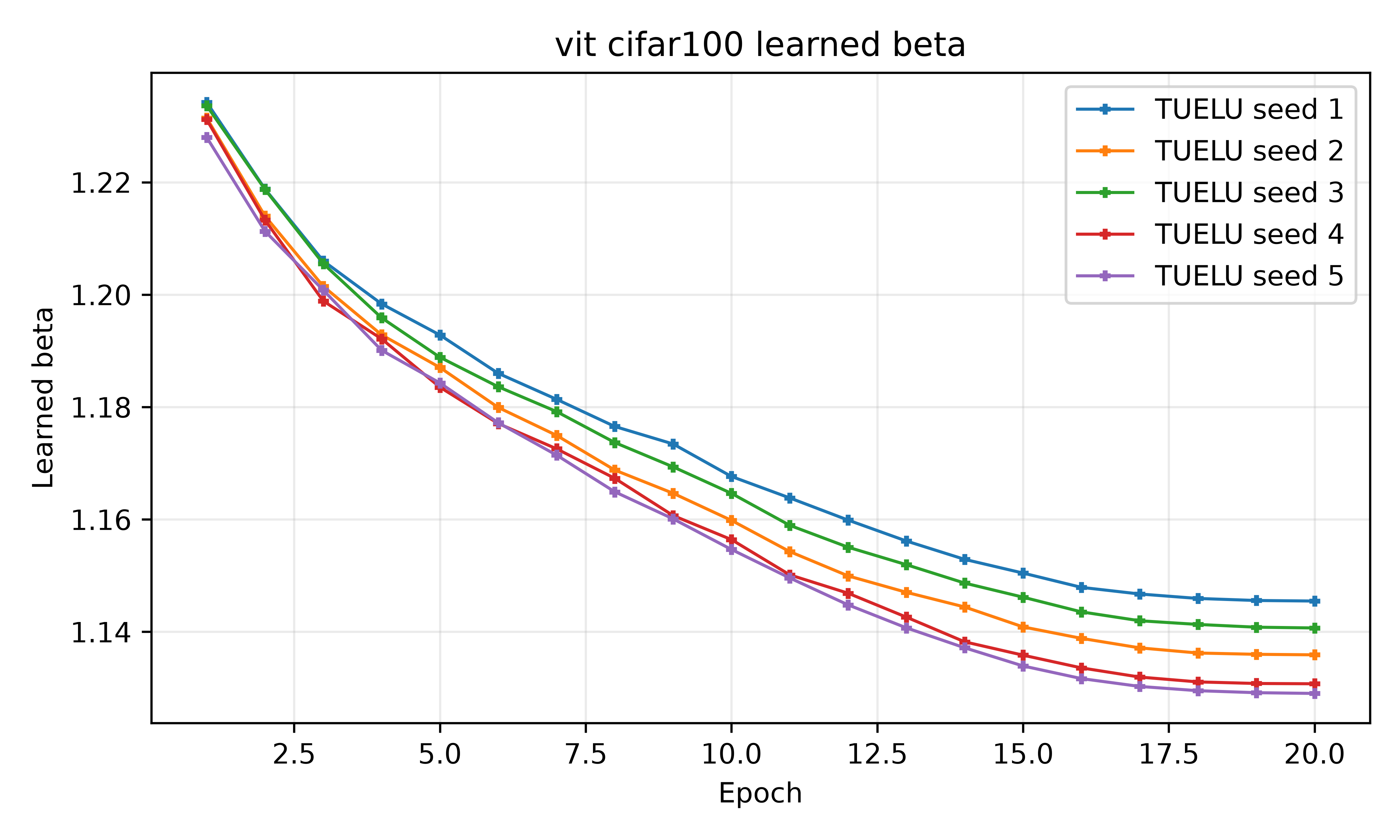}
\end{minipage}
\caption{Compact Vision Transformer results. Left: validation accuracy over training. Right: learned shared $\beta$ for TUELU.}
\label{fig:vit-results}
\end{figure*}

Table~\ref{tab:vit-results} and Figure~\ref{fig:vit-results} show that UELU with $\beta=1.25$ and TUELU improve over GELU, ReLU, and SiLU on mean test accuracy and test loss, whereas $\beta=3$ again underperforms. TUELU learns $\beta\approx1.14$, slightly narrower than the GELU-matched initialisation but still much closer to it than to the hard-swish width.

\begin{table*}[tbp]
\centering
\resizebox{\textwidth}{!}{
\begin{tabular}{lllllll}
\toprule
Activation & $\beta$ & Learned $\beta$ & Val. acc. & Test acc. & Test loss & Regions C/T/O \\
\midrule
ReLU & -- & -- & $46.70\pm0.51$ & $48.11\pm0.41$ & $2.4630\pm0.0178$ & -- \\
SiLU & -- & -- & $44.78\pm0.71$ & $46.35\pm0.60$ & $2.5171\pm0.0228$ & -- \\
GELU & -- & -- & $47.22\pm0.45$ & $48.80\pm0.54$ & $2.4458\pm0.0156$ & -- \\
DGELU & -- & -- & $44.34\pm0.61$ & $45.81\pm0.65$ & $2.5339\pm0.0248$ & -- \\
Hard swish / UELU & $3.0$ & -- & $44.10\pm0.62$ & $45.21\pm0.58$ & $2.5479\pm0.0174$ & $0.22/99.7/0.06$ \\
UELU & $1.25$ & -- & \best{48.09\pm0.40} & $49.28\pm0.33$ & \best{2.4281\pm0.0106} & $10.3/87.4/2.32$ \\
TUELU & $1.25$ & $1.136\pm0.006$ & $48.06\pm0.41$ & \best{49.41\pm0.34} & $2.4295\pm0.0099$ & $12.6/84.3/3.15$ \\
\bottomrule
\end{tabular}
}
\caption{Compact Vision Transformer on CIFAR-100. Validation and test accuracies, test loss, and learned $\beta$ are reported as mean $\pm$ standard deviation over five seeds. The final column records mean percentages of UELU preactivations in the closed, transition, and open regions.}
\label{tab:vit-results}
\end{table*}

\subsection{Tiny character-level GPT}

The third experiment uses a small decoder-only Transformer \cite{NIPS2017_3f5ee243} on Tiny Shakespeare,\footnote{\url{https://github.com/karpathy/char-rnn}} treated as character-level modelling with a 90\%/10\% train-validation split. The model has four Transformer blocks, four attention heads, embedding dimension 128, context length 128, dropout 0.1, and the compared activation in the feedforward sublayers. Training uses AdamW with learning rate $3\cdot10^{-4}$, minimum learning rate $3\cdot10^{-5}$, weight decay $0.1$, gradient clipping at 1.0, batch size 64, 100 warmup iterations, cosine decay, and 10,000 final-comparison iterations.

\begin{figure*}
\centering
\begin{minipage}{0.48\textwidth}
\centering
\includegraphics[width=\textwidth]{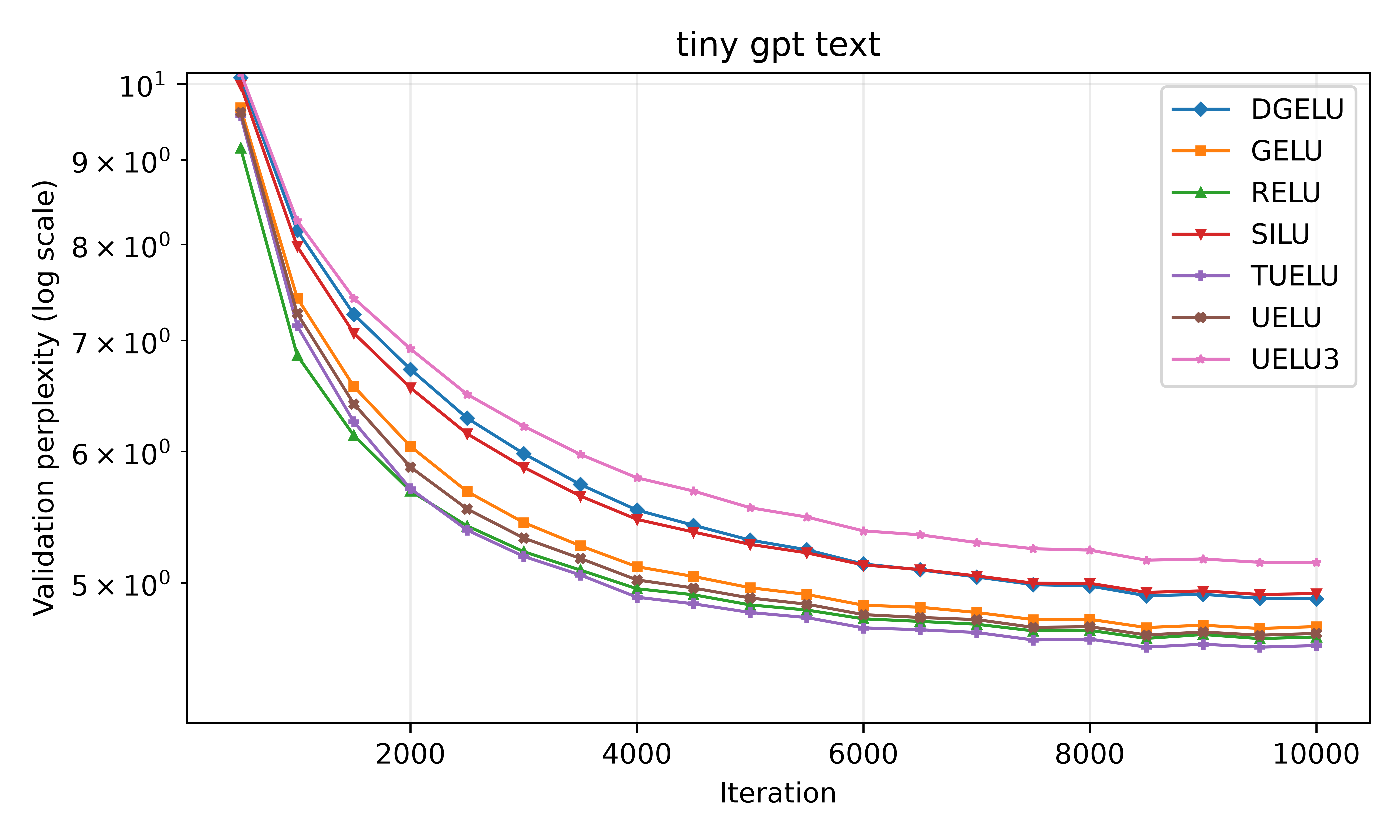}
\end{minipage}\hfill
\begin{minipage}{0.48\textwidth}
\centering
\includegraphics[width=\textwidth]{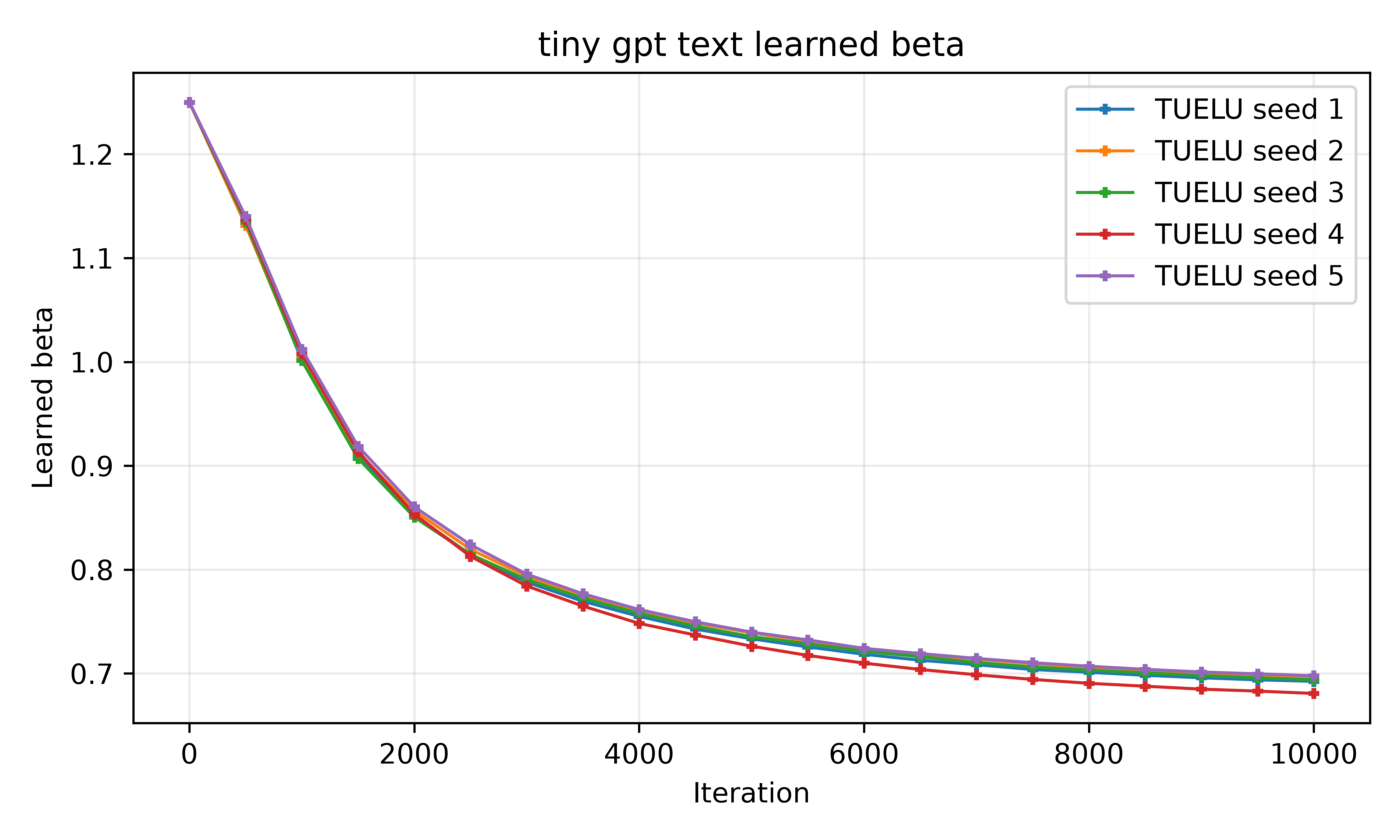}
\end{minipage}
\caption{Tiny GPT results. Left: validation perplexity over training. Right: learned shared $\beta$ for TUELU.}
\label{fig:gpt-results}
\end{figure*}

Table~\ref{tab:gpt-results} and Figure~\ref{fig:gpt-results} show that TUELU improves over all baselines, reducing mean validation perplexity to $4.575$. Fixed UELU with $\beta=1.25$ improves over GELU and SiLU, but is slightly behind ReLU in this character-level setting. The $\beta=3$ setting again performs poorly, while TUELU learns a smaller width, $\beta\approx0.70$.

\begin{table*}[tbp]
\centering
\resizebox{\textwidth}{!}{
\begin{tabular}{llllll}
\toprule
Activation & $\beta$ & Learned $\beta$ & Val. loss & Perplexity & Regions C/T/O \\
\midrule
ReLU & -- & -- & $1.5317\pm0.0058$ & $4.626\pm0.027$ & -- \\
SiLU & -- & -- & $1.5920\pm0.0085$ & $4.914\pm0.042$ & -- \\
GELU & -- & -- & $1.5475\pm0.0087$ & $4.700\pm0.041$ & -- \\
DGELU & -- & -- & $1.5848\pm0.0080$ & $4.879\pm0.039$ & -- \\
Hard swish / UELU & $3.0$ & -- & $1.6357\pm0.0079$ & $5.133\pm0.041$ & $0.10/99.9/0.01$ \\
UELU & $1.25$ & -- & $1.5383\pm0.0081$ & $4.657\pm0.038$ & $4.44/95.4/0.18$ \\
TUELU & $1.25$ & $0.696\pm0.007$ & \best{1.5205\pm0.0081} & \best{4.575\pm0.037} & $10.9/87.9/1.18$ \\
\bottomrule
\end{tabular}
}
\caption{Tiny character-level GPT. Validation loss, perplexity, and learned $\beta$ are reported as mean $\pm$ standard deviation over five seeds. The final column records mean percentages of UELU preactivations in the closed, transition, and open regions.}
\label{tab:gpt-results}
\end{table*}

\subsection{TinyStories token-level GPT}

The fourth experiment keeps the decoder-only Transformer setting but moves from character-level Tiny Shakespeare to token-level TinyStories \cite{DBLP:journals/corr/abs-2305-07759}. We train a byte-level BPE tokenizer \cite{sennrich-etal-2016-neural} with 8,000 tokens on the training split, use up to 50,000 training stories and 5,000 validation stories, and compare activations in the feedforward sublayers. The model has six Transformer blocks, six attention heads, embedding dimension 384, context length 256, dropout 0.1, and batch size 32. 
The optimiser and schedule match the Tiny character-level GPT experiment.

\begin{figure*}
\centering
\begin{minipage}{0.48\textwidth}
\centering
\includegraphics[width=\textwidth]{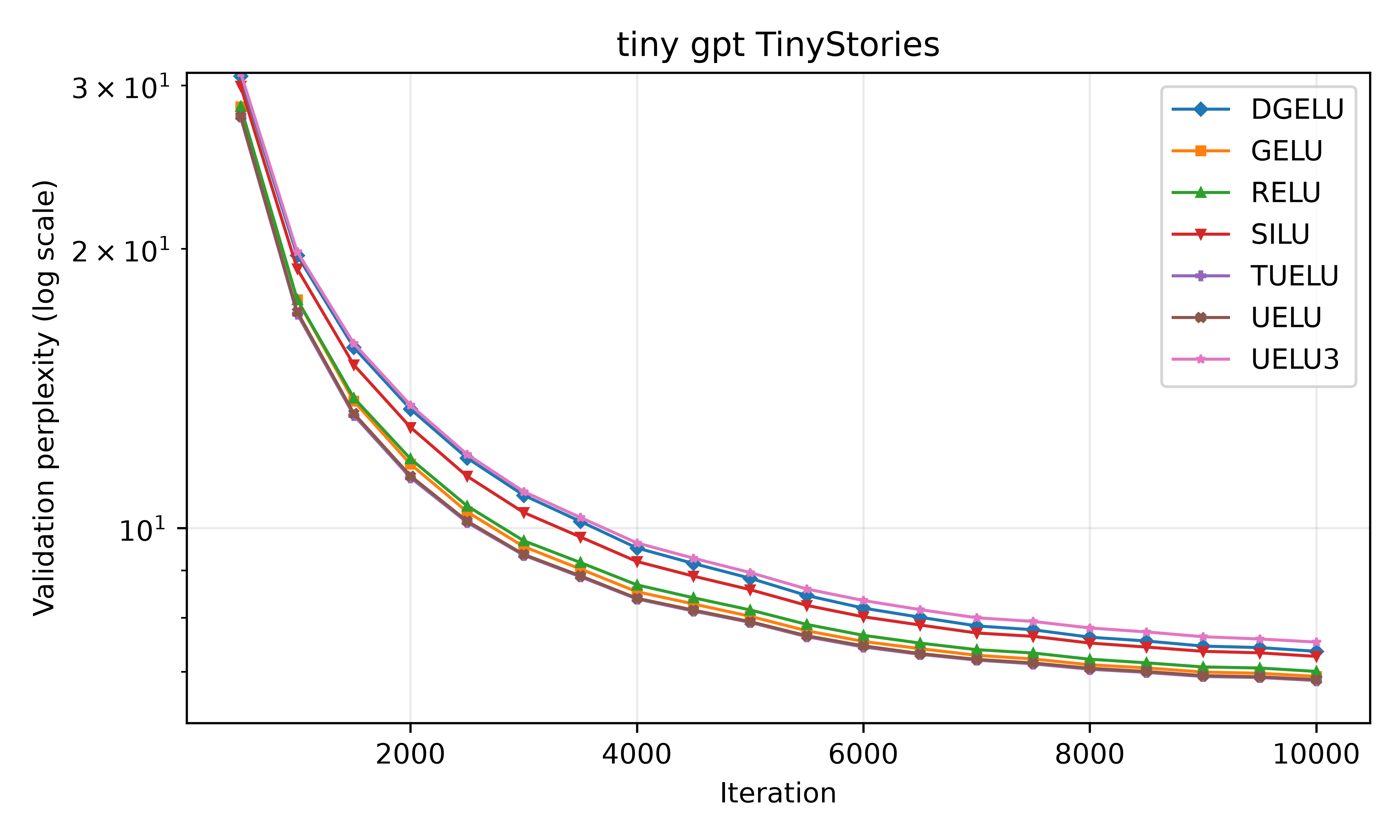}
\end{minipage}\hfill
\begin{minipage}{0.48\textwidth}
\centering
\includegraphics[width=\textwidth]{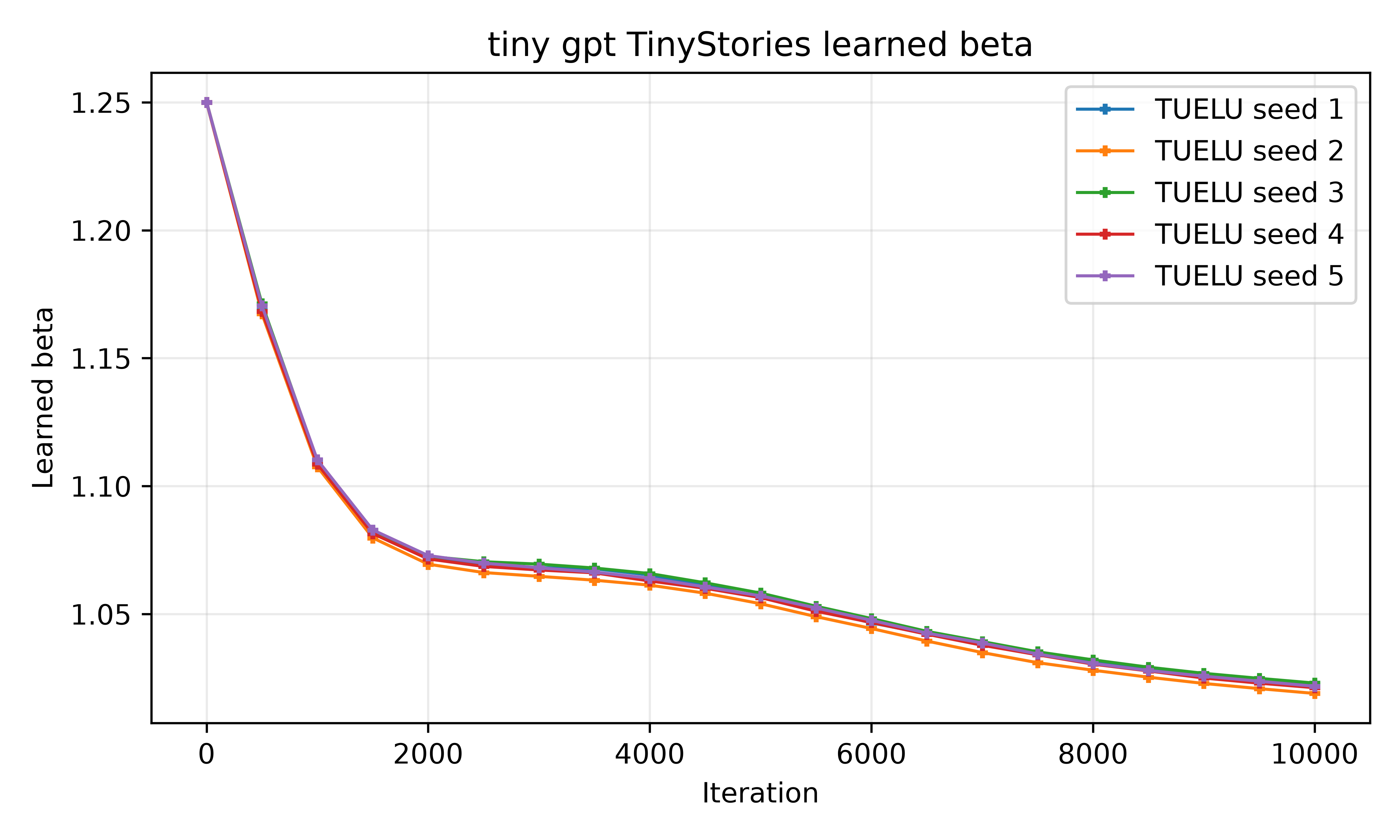}
\end{minipage}
\caption{TinyStories GPT results. Left: validation perplexity over training. Right: learned shared $\beta$ for TUELU.}
\label{fig:tinystories-gpt-results}
\end{figure*}

Table~\ref{tab:tinystories-gpt-results} and Figure~\ref{fig:tinystories-gpt-results} show that fixed UELU with $\beta=1.25$ and TUELU improve over GELU, with mean validation perplexities $6.879$ and $6.863$ versus $6.935$. TUELU learns $\beta\approx1.02$, narrower than the GELU-matched initialisation but wider than in the character-level setting. The hard-swish-width $\beta=3$ setting again performs poorly.

\begin{table*}[tbp]
\centering
\resizebox{\textwidth}{!}{
\begin{tabular}{llllll}
\toprule
Activation & $\beta$ & Learned $\beta$ & Val. loss & Perplexity & Regions C/T/O \\
\midrule
ReLU & -- & -- & $1.9498\pm0.0065$ & $7.028\pm0.046$ & -- \\
SiLU & -- & -- & $1.9865\pm0.0076$ & $7.290\pm0.055$ & -- \\
GELU & -- & -- & $1.9366\pm0.0078$ & $6.935\pm0.054$ & -- \\
DGELU & -- & -- & $1.9999\pm0.0089$ & $7.389\pm0.066$ & -- \\
Hard swish / UELU & $3.0$ & -- & $2.0224\pm0.0068$ & $7.556\pm0.051$ & $2.82/97.2/0.02$ \\
UELU & $1.25$ & -- & $1.9284\pm0.0070$ & $6.879\pm0.048$ & $20.2/79.4/0.35$ \\
TUELU & $1.25$ & $1.021\pm0.001$ & \best{1.9261\pm0.0070} & \best{6.863\pm0.048} & $25.7/73.7/0.66$ \\
\bottomrule
\end{tabular}
}
\caption{TinyStories token-level GPT. Validation loss, perplexity, and learned $\beta$ are reported as mean $\pm$ standard deviation over five seeds. The final column records mean percentages of UELU preactivations in the closed, transition, and open regions.}
\label{tab:tinystories-gpt-results}
\end{table*}

\subsection{WikiText-2 token-level GPT}

The fifth experiment repeats the token-level GPT protocol on WikiText-2 \cite{merity2016pointer}. This provides a conventional language-modelling benchmark based on encyclopedia-style text, complementing the simpler story-like distribution of TinyStories. We use the WikiText-2 raw train and validation splits, train the same 8,000-token byte-level BPE tokenizer family, and compare activations in the feedforward sublayers of a small decoder-only Transformer. The model has four Transformer blocks, four attention heads, embedding dimension 256, context length 256, dropout 0.1, and batch size 32. 
The optimiser and schedule match the Tiny character-level GPT experiment.

\begin{table*}[tbp]
\centering
\resizebox{\textwidth}{!}{
\begin{tabular}{llllll}
\toprule
Activation & $\beta$ & Learned $\beta$ & Val. loss & Perplexity & Regions C/T/O \\
\midrule
ReLU & -- & -- & $4.4345\pm0.0101$ & $84.312\pm0.854$ & -- \\
SiLU & -- & -- & $4.4640\pm0.0096$ & $86.835\pm0.836$ & -- \\
GELU & -- & -- & $4.4172\pm0.0154$ & $82.875\pm1.276$ & -- \\
DGELU & -- & -- & $4.4696\pm0.0099$ & $87.329\pm0.868$ & -- \\
UELU & $1.25$ & -- & $4.4166\pm0.0155$ & $82.824\pm1.276$ & $11.7/88.1/0.18$ \\
Hard swish / UELU & $3.0$ & -- & $4.5092\pm0.0087$ & $90.849\pm0.793$ & $0.45/99.5/0.02$ \\
TUELU & $1.25$ & $0.887\pm0.003$ & \best{4.4165\pm0.0145} & \best{82.816\pm1.195} & $18.1/81.3/0.59$ \\
\bottomrule
\end{tabular}
}
\caption{WikiText-2 token-level GPT. Validation loss, perplexity, and learned $\beta$ are reported as mean $\pm$ standard deviation over five seeds. The final column records mean percentages of UELU preactivations in the closed, transition, and open regions.}
\label{tab:wikitext2-gpt-results}
\end{table*}

\begin{figure*}
\centering
\begin{minipage}{0.48\textwidth}
\centering
\includegraphics[width=\textwidth]{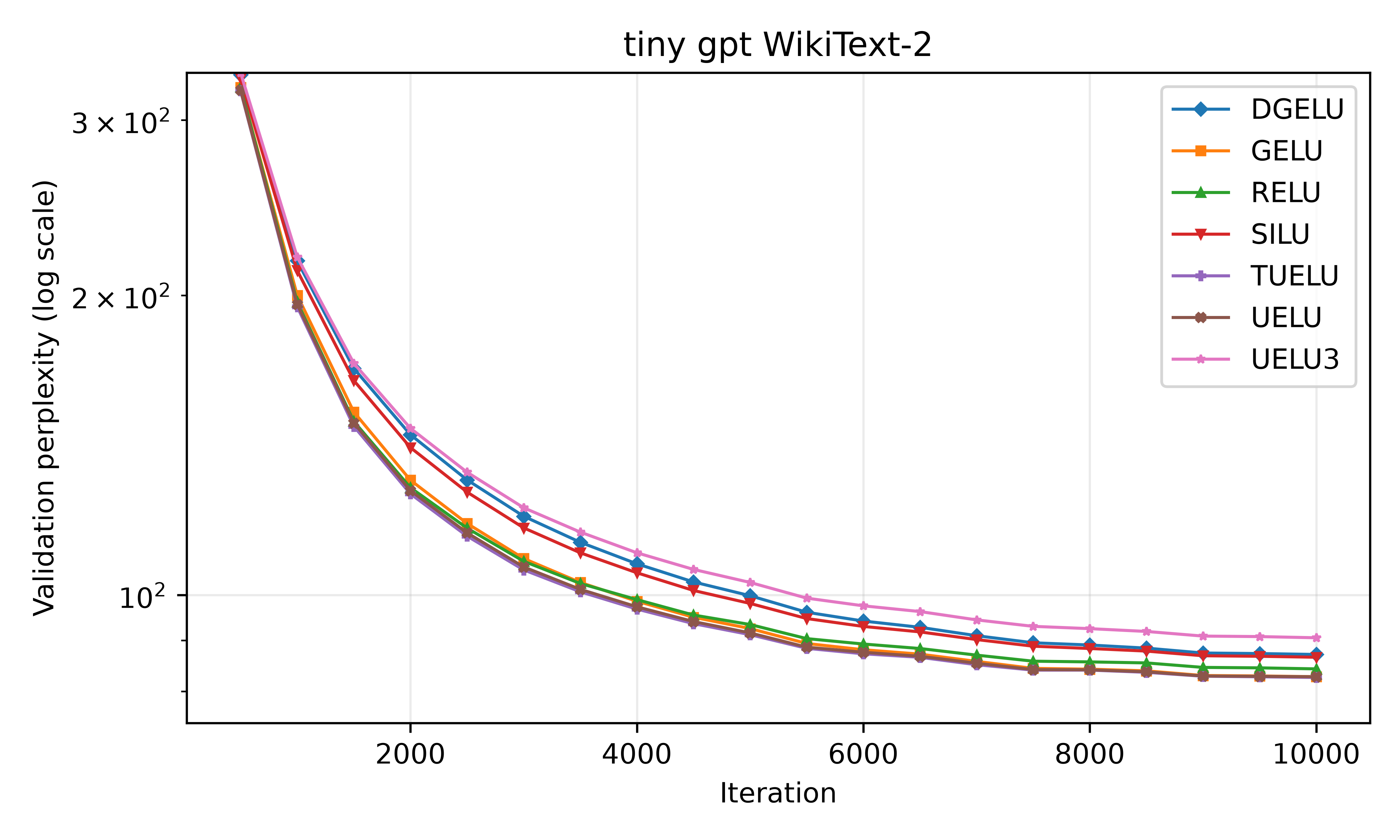}
\end{minipage}\hfill
\begin{minipage}{0.48\textwidth}
\centering
\includegraphics[width=\textwidth]{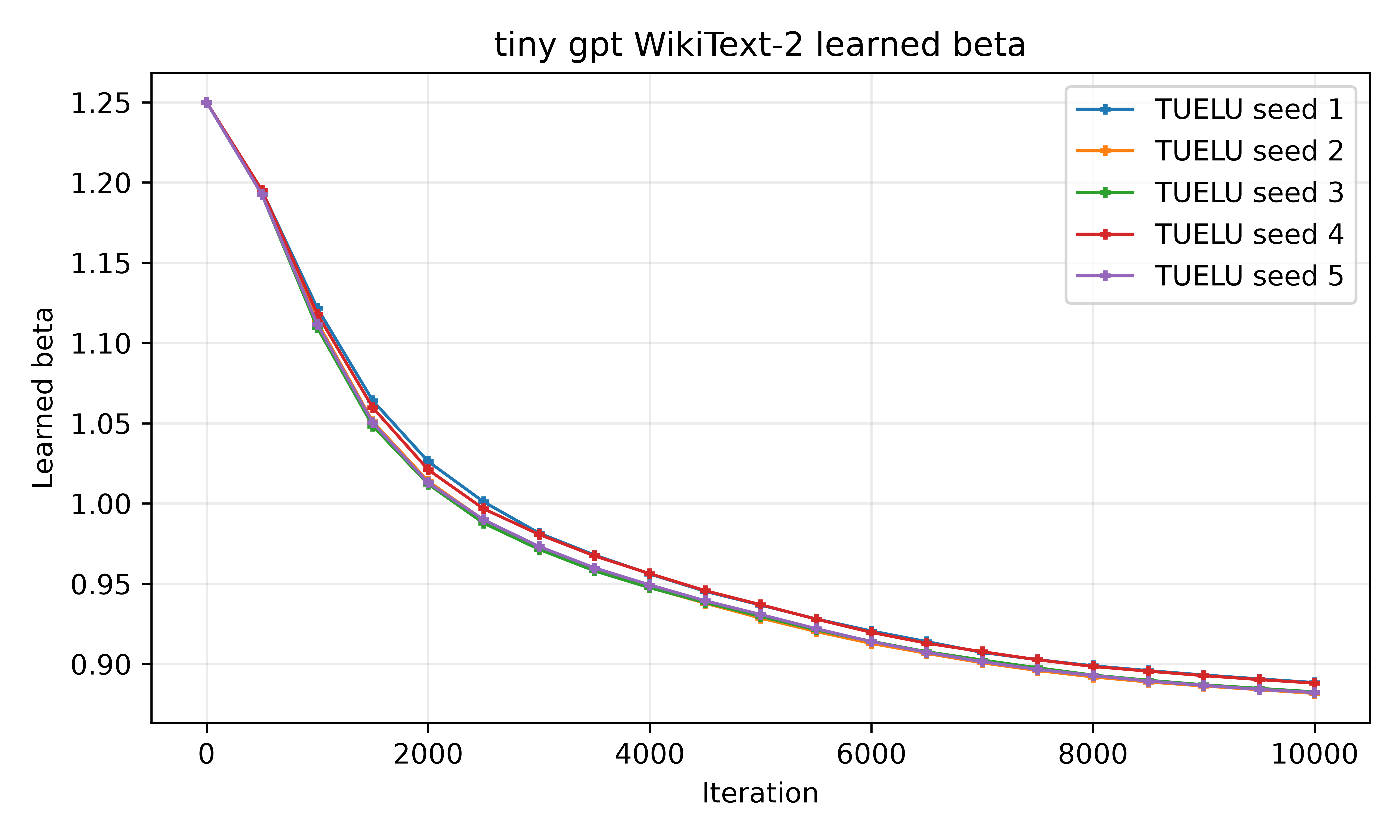}
\end{minipage}
\caption{WikiText-2 GPT results. Left: validation perplexity over training. Right: learned shared $\beta$ for TUELU.}
\label{fig:wikitext2-gpt-results}
\end{figure*}

Table~\ref{tab:wikitext2-gpt-results} and Figure~\ref{fig:wikitext2-gpt-results} show that GELU, UELU, and TUELU are essentially tied on this benchmark, with TUELU giving the lowest mean validation perplexity at $82.816$ and fixed UELU reaching $82.824$ versus $82.875$ for GELU. TUELU learns a narrower threshold, $\beta\approx0.89$. The conventional hard-swish-width setting $\beta=3$ again performs poorly.

\subsection{Region occupancy}

\begin{figure}
\centering
\includegraphics[width=0.65\columnwidth]{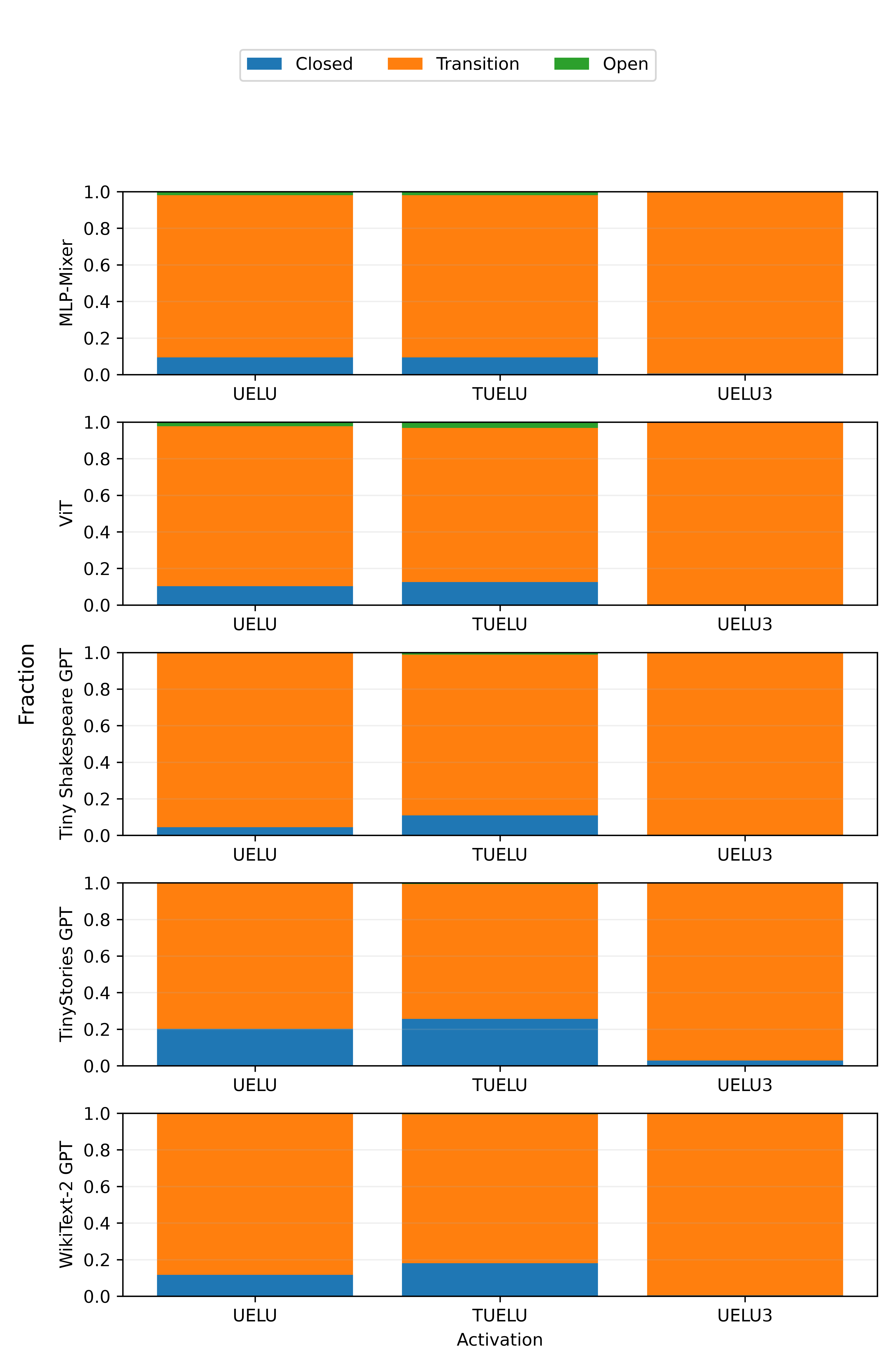}
\caption{Closed, transition, and open region occupancy for fixed-width UELU and TUELU across the computational study. Percentages are computed from held-out validation batches.}
\label{fig:occupancy-results}
\end{figure}

In this section we discuss learned width and region occupancy for the uniform-threshold members. The region occupancy summaries are collected in Figure~\ref{fig:occupancy-results}. GELU-matched UELU and TUELU produce nontrivial closed and open regions, while $\beta=3$ (i.e. hard swish) places almost all observed preactivations in the transition region. The learned TUELU widths are architecture dependent: approximately $1.25$ for Mixer, $1.14$ for Vision Transformer, $0.70$ for character-level GPT, $1.02$ for TinyStories GPT, and $0.89$ for WikiText-2 GPT. Thus the learned widths stay closer to the GELU-matched scale than to the conventional hard-swish width, while still adapting downward in the language models. The learned trajectories are consistent across seeds. Across these controlled settings, fixed or learned UELU is competitive with standard activations; DGELU is consistently weaker, suggesting that the first-order-loss correction is conceptually informative, but retaining a remnant of it does not appear useful during training.

\section{Discussion and conclusion}
\label{sec:discussion-conclusion}

This work identifies GELU as the signal-transmission component of the Gaussian complementary first-order loss function. The full loss measures expected surplus and includes the truncated-moment correction $\phi(z)$; GELU keeps only $z\Phi(z)$, the probability-weighted signal transmitted across a Gaussian uncertain threshold. This explains why the full complementary loss and its centred version have different activation geometry: one fails to preserve the origin, while the other creates a negative plateau. Removing $\phi(z)$ is therefore the step that converts a loss-accounting quantity into an origin-preserving signal-transmission operator.

The same decomposition gives a threshold-transmission family $zF(z)$. ReLU, GELU, SiLU/Swish, UELU/hard swish, and ExpELU correspond to different threshold laws. In the uniform case, the half-width $\beta$ has a direct interpretation as the size of the uncertain boundary layer, giving both fixed-width UELU and learned-width TUELU.

In the controlled experiments, GELU-matched or learned uniform-threshold gates are consistently competitive with standard activations and use the finite transition region nontrivially. The learned widths differ by architecture but remain far below the conventional hard-swish setting $\beta=3$, which is consistently too broad and places almost all observed preactivations in the transition region. 

The empirical study is deliberately controlled and small in scale. 
Broader benchmarking, additional adaptive activation baselines such as full AQuLU, per-layer or per-channel threshold widths, and larger-scale vision and language experiments are left for future work. 
Finally, we do not investigate runtime advantages: although UELU has a simple clipped piecewise-polynomial form --- which is computationally appealing --- practical speed depends on backend kernels, operator fusion, and hardware. Modern activation function implementations are typically highly optimised, therefore this investigation is not trivial and is left for future work. 

\paragraph{Code availability} The supporting code used to reproduce the experiments in this work is available at \url{https://github.com/gwr3n/uelu}.

\appendix

\section{Derivations for logistic, exponential \& uniform threshold laws}
\label{app:alternative-activations}

This appendix gives the partial-moment decompositions underlying SiLU/Swish, ExpELU, and UELU. The derivations parallel the Gaussian calculation in the main text.

\subsection{Logistic threshold}

Let $ T $ have the standard logistic distribution. Then
$$
F(t)=\sigma(t)=\frac{1}{1+e^{-t}},
\qquad
f(t)=\sigma(t)(1-\sigma(t)).
$$
The complementary first-order loss is
$$
\widehat{L}_{\log}(z)
=
\mathbb{E}[(z-T)^+]
=
\int_{-\infty}^{z}(z-t)f(t)\,dt.
$$
Expanding the integral gives
$$
\widehat{L}_{\log}(z)
=
z\int_{-\infty}^{z}f(t)\,dt
-
\int_{-\infty}^{z}t f(t)\,dt.
$$
The first integral is
$$
\int_{-\infty}^{z}f(t)\,dt
=
F(z)
=
\sigma(z).
$$
For the second integral, use $f(t)=\sigma'(t)$. Integration by parts gives
$$
\int_{-\infty}^{z}t f(t)\,dt
=
\int_{-\infty}^{z}t\sigma'(t)\,dt
=
z\sigma(z)-\int_{-\infty}^{z}\sigma(t)\,dt.
$$
Since
$$
\int_{-\infty}^{z}\sigma(t)\,dt
=
\left[\log(1+e^t)\right]_{-\infty}^{z}
=
\log(1+e^z),
$$ 
we obtain
$$
\int_{-\infty}^{z}t f(t)\,dt
=
z\sigma(z)-\log(1+e^z).
$$
Hence
$$
\widehat{L}_{\log}(z)
=
z\sigma(z)
-
\bigl(z\sigma(z)-\log(1+e^z)\bigr)
=
\log(1+e^z).
$$
Thus the logistic complementary first-order loss is the softplus function. Written as signal transmission plus a truncated-moment correction, this is
$$\widehat{L}_{\log}(z)=\underbrace{z\sigma(z)}_{\text{boundary-gated signal}}+\quad\underbrace{\bigl(\log(1+e^z)-z\sigma(z)\bigr)}_{\text{truncated-moment correction}}.$$
Discarding the correction yields the SiLU/Swish activation, $z\sigma(z)$.

\subsection{Exponential threshold}

Let $T$ have an exponential distribution with rate $\lambda>0$:
$$
f(t)=\lambda e^{-\lambda t},\quad
F(t)=1-e^{-\lambda t},\qquad t\ge 0.
$$
For $ z\le 0 $, the complementary first-order loss is zero. For $ z>0 $,
$$
\widehat{L}_{\exp}(z)
=
\mathbb{E}[(z-T)^+]
=
\int_0^z (z-t)\lambda e^{-\lambda t}\,dt.
$$
Using $\int_0^z \lambda e^{-\lambda t}\,dt=1-e^{-\lambda z}$ and $\int_0^z t\lambda e^{-\lambda t}\,dt=-ze^{-\lambda z}+\tfrac{1}{\lambda}(1-e^{-\lambda z})$,
$$
\widehat{L}_{\exp}(z)
=
z-\frac{1}{\lambda}(1-e^{-\lambda z}).
$$Written as signal transmission minus truncated-moment accounting, the same expression is
$$
\widehat{L}_{\exp}(z)
=
\underbrace{z(1-e^{-\lambda z})}_{\text{boundary-gated signal}}
-\quad
\underbrace{\left(\frac{1}{\lambda}(1-e^{-\lambda z})-ze^{-\lambda z}\right)}_{\text{truncated-moment correction}}.
$$
Discarding the correction yields
$$
\operatorname{ExpELU}_{\lambda}(z)
=
\begin{cases}
0, & z\le 0,\\
z(1-e^{-\lambda z}), & z>0.
\end{cases}
$$

\subsection{Uniform threshold}

Let $ T\sim U(-\beta,\beta) $, where $ \beta>0 $. Then
$$
F(t)=
\begin{cases}
0, & t<-\beta,\\
\dfrac{t+\beta}{2\beta}, & -\beta\le t\le \beta,\\
1, & t>\beta.
\end{cases}
$$
The complementary first-order loss is zero for $ z<-\beta $. For $ -\beta\le z\le \beta $,
$$
\widehat{L}_{U}(z)
=
\mathbb{E}[(z-T)^+]
=
\int_{-\beta}^z (z-t)\frac{1}{2\beta}\,dt.
$$
Therefore
$$
\widehat{L}_{U}(z)
=
\frac{(z+\beta)^2}{4\beta}.
$$
Since $ F(z)=(z+\beta)/(2\beta) $ in this region, 
$$
\widehat{L}_{U}(z)
=
\underbrace{\frac{z(z+\beta)}{2\beta}}_{\text{boundary-gated signal}}
-
\underbrace{\frac{z^2-\beta^2}{4\beta}}_{\text{truncated-moment correction}}.
$$
For $ z>\beta $,
$$
\widehat{L}_{U}(z)
=
\int_{-\beta}^{\beta} (z-t)\frac{1}{2\beta}\,dt
=
z.
$$
In this region, $ F(z)=1 $, and the mean of the symmetric uniform threshold is zero. Therefore the complementary first-order loss and the boundary-gated signal coincide:
$$
\widehat{L}_{U}(z)
=
\underbrace{z}_{\text{boundary-gated signal}}
-
\underbrace{0}_{\text{truncated-moment correction}}.
$$
Discarding the correction in each region gives
$$
\operatorname{UELU}_{\beta}(z)
=
zF(z)
=
\begin{cases}
0, & z<-\beta,\\
\dfrac{z(z+\beta)}{2\beta}, & -\beta\le z\le \beta,\\
z, & z>\beta.
\end{cases}
$$

\subsection{Universal approximation}

The threshold-transmission activations considered satisfy the standard non-polynomial condition for universal approximation. By the theorem of Leshno et al.  \cite{Leshno1993}, finite sums
$\sum_{j=1}^{m} a_j \phi(w_j^\top x+b_j)$
are dense in $C(K)$ --- the space of all continuous real-valued functions on the set $K$ --- for every compact $K\subset\mathbb{R}^n$ whenever $\phi$ is locally bounded, piecewise continuous, and not equal almost everywhere to an algebraic polynomial.

For $\lambda>0$, $\operatorname{ExpELU}_{\lambda}$ is non-polynomial because on $(0,\infty)$ it equals $z-ze^{-\lambda z}$. For $\beta>0$, $\operatorname{UELU}_{\beta}$ is also not equal almost everywhere to any single polynomial: it is zero on $(-\infty,-\beta)$, quadratic on $(-\beta,\beta)$, and linear on $(\beta,\infty)$. A polynomial agreeing with it on the open interval $(-\infty,-\beta)$ would have to be zero, which cannot also agree with $z$ on $(\beta,\infty)$. Thus fixed-width UELU networks retain the usual universal approximation property. The same conclusion applies to TUELU for every finite learned width $\beta>0$.

\section{Table of symbols}
\label{app:symbols}

In this section we provide Table \ref{tab:gelu-statement-decomposition}, which is intended to support readers navigating the notation and terminology used in our derivations.

\begin{table*}[tbp]
\centering
\small
\begin{adjustbox}{max width=\textwidth}
\begin{tabular}{@{}lll@{}}
\toprule
Phrase & Mathematical object & Meaning \\
\midrule
Input signal
&
$z$
&
The preactivation value to be transmitted or suppressed.
\\
Gaussian random threshold
&
$T\sim N(0,1)$
&
The uncertain boundary that the input must exceed.
\\
Hard threshold event
&
$\{T\le z\}$
&
The event that the input clears the threshold.
\\
Hard gate indicator
&
$\mathbf{1}_{\{T\le z\}}$
&
Equals one when the gate is open and zero otherwise.
\\
Hard linear gate
&
$z\mathbf{1}_{\{T\le z\}}$
&
Transmits $z$ unchanged if the threshold is cleared, and 0 otherwise.
\\
Gate-opening probability
&
$\mathbb{P}(T\le z)=\Phi(z)$
&
The probability that the Gaussian threshold lies below the input.
\\
Expected transmitted signal
&
$\mathbb{E}\!\left[z\mathbf{1}_{\{T\le z\}}\right]$
&
The average output of the hard linear gate.
\\
Signal-transmission term
&
$z\mathbb{P}(T\le z)=z\Phi(z)$
&
The input signal multiplied by its probability of transmission.
\\
Hard surplus gate
&
$(z-T)^+=(z-T)\mathbf{1}_{\{T\le z\}}$
&
Measures the positive excess of the input over the threshold.
\\
Expected surplus
&
$\mathbb{E}[(z-T)^+]$
&
The expected amount by which $z$ exceeds the random threshold.
\\
Gaussian complementary first-order loss
&
$\widehat{L}(z)=\mathbb{E}[(z-T)^+]$
&
The expected-surplus function associated with a Gaussian threshold.
\\
Truncated-moment correction
&
$-\mathbb{E}\!\left[T\mathbf{1}_{\{T\le z\}}\right]=\phi(z)$
&
The loss-accounting term needed to compute expected surplus.
\\
GELU
&
$\operatorname{GELU}(z)=z\Phi(z)$
&
The signal-transmission term retained as the activation.
\\
\bottomrule
\end{tabular}
\end{adjustbox}
\caption{Table of mathematical objects utilised in our structural interpretation of GELU.}
\label{tab:gelu-statement-decomposition}
\end{table*}

\bibliographystyle{plain}
\bibliography{references}

\end{document}